\documentclass[conference]{IEEEtran}
\pdfoutput=1
\IEEEoverridecommandlockouts
\usepackage{cite}
\usepackage{amsmath,amssymb,amsfonts}
\usepackage{graphicx}
\usepackage{textcomp}
\usepackage[dvipsnames]{xcolor}
\usepackage{hyperref}
\usepackage{multirow}
\hypersetup{
    colorlinks=true,
    linkcolor=blue,
    filecolor=magenta,      
    urlcolor=blue,
}
\usepackage{soul}
\usepackage{booktabs}

\usepackage{algorithm}
\usepackage[algo2e]{algorithm2e} 
\usepackage{algpseudocode}
\usepackage{comment}
\usepackage{multirow}
\usepackage{float}
\usepackage[caption=false]{subfig}
\usepackage{framed}
\usepackage[normalem]{ulem}
\def\BibTeX{{\rm B\kern-.05em{\sc i\kern-.025em b}\kern-.08em
    T\kern-.1667em\lower.7ex\hbox{E}\kern-.125emX}}

\title{Real-Time Human Action Recognition on Embedded Platforms}
\author{
\IEEEauthorblockN{Ruiqi Wang\IEEEauthorrefmark{1}\IEEEauthorrefmark{2}, Zichen Wang\IEEEauthorrefmark{1}\IEEEauthorrefmark{2}, Peiqi Gao\IEEEauthorrefmark{1}\IEEEauthorrefmark{2}, Mingzhen Li\IEEEauthorrefmark{1}\IEEEauthorrefmark{2}, Jaehwan Jeong\IEEEauthorrefmark{1}\IEEEauthorrefmark{2}, \\Yihang Xu\IEEEauthorrefmark{1}\IEEEauthorrefmark{2}, Yejin Lee\IEEEauthorrefmark{1}\IEEEauthorrefmark{3}, Carolyn M. Baum\IEEEauthorrefmark{1}\IEEEauthorrefmark{3}, Lisa Tabor Connor\IEEEauthorrefmark{1}\IEEEauthorrefmark{3}, Chenyang Lu\IEEEauthorrefmark{1}\IEEEauthorrefmark{2}}
\IEEEauthorblockA{
\IEEEauthorrefmark{1}\textit{AI for Health Institute, Washington University in St. Louis}, \\
\IEEEauthorrefmark{2}\textit{Department of Computer Science and Engineering, Washington University in St. Louis}, \\
\IEEEauthorrefmark{3}\textit{Program in Occupational Therapy, Washington University School of Medicine}
\\
\{ruiqi.w, zichen.wang, peiqi, l.mingzhen, j.jaehwan, x.yihang, yejin.l, baumc, lconnor, lu\}@wustl.edu}

}
\newcommand {\ie} {{i.e., }}
\newcommand {\eg} {{e.g., }}

\newcommand {\beq} {\begin{equation}}
\newcommand {\eeq} {\end{equation}}
\newcommand {\bequn} {\begin{equation*}}
\newcommand {\eequn} {\end{equation*}}
\newcommand {\bear} {\begin{eqnarray}}
\newcommand {\eear} {\end{eqnarray}}
\newcommand {\bearun} {\begin{eqnarray*}}
\newcommand {\eearun} {\end{eqnarray*}}

\newcommand{\proposed}{{IMFE}\xspace}
\newcommand{\framework}{{RT-HARE}\xspace}
\begin{document}
\maketitle
\thispagestyle{plain}
\pagestyle{plain}

\begin{abstract}

With advancements in computer vision and deep learning, video-based human action recognition (HAR) has become practical. However,  due to the complexity of the computation pipeline, running HAR on live video streams incurs excessive delays on embedded platforms. This work tackles the real-time performance challenges of HAR with four contributions: 1) an experimental study identifying a standard Optical Flow (OF) extraction technique as the latency bottleneck in a state-of-the-art HAR pipeline, 2) an exploration of the latency-accuracy tradeoff between the standard and deep learning approaches to OF extraction, which highlights the need for a novel, efficient motion feature extractor, 3) the design of Integrated Motion Feature Extractor (IMFE), a novel single-shot neural network architecture for motion feature extraction with drastic improvement in latency, 4) the development of RT-HARE, a real-time HAR system tailored for embedded platforms. Experimental results on an Nvidia Jetson Xavier NX platform demonstrated that RT-HARE realizes real-time HAR at a video frame rate of 30 frames per second while delivering high levels of recognition accuracy.

\end{abstract}


\section{Introduction}
\label{sec:intro}
As video-based human action recognition (HAR) technology became widely available in recent years~\cite{li2020mstcn, asformer}, we have witnessed the emergence of applications requiring HAR to be performed on real-time using live video streams. For example, 
HAR has been employed to monitor the activities of Alzheimer's disease patients and provides alerts when anomalies are detected \cite{snoun2023deep}. HAR has also been used to interpret dangerous driver behaviors~\cite{10208426} for just-in-time interventions to improve driving safety. Moreover, due to privacy concerns, it is often preferable to deploy HAR on local embedded platforms instead of relying on servers or the cloud.

However, achieving real-time performance in HAR remains extremely challenging on embedded platforms due to the complexity of the computational pipelines and the resource constraints inherent to these platforms. Strikingly, in our benchmark experiments, a state-of-the-art HAR pipeline~\cite{xu2021long} can handle only three video frames per second (FPS) due to its excessive and fluctuating latency.

We first implemented and benchmarked a state-of-the-art two-stream HAR architecture~\cite{simonyan2014two} on a Nvidia Jetson platform to understand real-time performance issues. The HAR pipeline employs two feature streams extracted from video frames: an RGB stream encoding the spatial information from static video frames and a motion stream encoding temporal information from sequences of optical flows (OFs) across video frames. A deep learning model then recognizes actions using the two feature streams. Surprisingly, our experimental study revealed the latency of the pipeline is dominated by the OF feature extractor using a standard OpenCV implementation. Furthermore, while a deep-learning-based OF extractor effectively reduces the latency, it incurs a substantial drop in recognition accuracy. The tradeoff between the standard and deep learning approaches to OF extraction highlights the need for a novel motion feature extractor design that is both efficient and accurate. 

To this end, we propose the Integrated Motion Feature Extractor (\proposed), a novel motion feature extractor specifically designed to meet these stringent requirements on embedded systems. In contrast to the existing motion feature extraction approach relying on computationally expensive OF extraction, \proposed introduces a lightweight single-shot motion feature extractor that directly generates motion features without the need for OF extraction. 

Based on \proposed, we have developed and evaluated \framework, an end-to-end HAR framework for real-time HAR on embedded platforms. Experiments on an Nvidia Jetson Xavier NX platform demonstrated that \framework can perform HAR at 30 video frames per second (10x improvement over the original HAR pipeline) while incurring only a moderate loss in recognition accuracy when compared to a server-based HAR system. While this work is motivated by the challenges in HAR, \proposed may be applied to any real-time video analytics applications using motion features, enhancing their efficiency and real-time performance on embedded platforms.

Specifically, our contributions are as follows:

\begin{itemize}
    \item Experimental study identifying a standard OF extractor as the latency bottleneck in a state-of-the-art HAR approach, 
    \item Exploration of the latency--accuracy tradeoff between traditional and deep learning approaches to OF extraction,
    \item Design of \proposed, a novel single-shot neural network architecture for motion feature extraction with both efficiency and accuracy,
    \item Implementation of \framework, a real-time HAR system optimized for embedded platforms. 
\end{itemize}

\section{Problem Formulation}
This section provides an overview of the problem statement of HAR and the soft real-time requirements inherent in such systems.

\subsection{Objective of HAR}
The objective of video-based HAR is to minimize the error between the recognized and true action labels for a set of videos. Given a video \( V \) consisting of a sequence of $M$ frames, $\{x_1, x_2, x_3, \cdots x_M \}$, the video frames are grouped into mini clips of length $K$, $\{X_1, X_2, \cdots, X_N\}$, where \( N = M/K\) is the total number of clips. A feature extractor $E$ extracts the video features $f$ from clips, $f_t = E(X_t)$, where $t \in {1, 2, ..., N}$ Finally, the video is represented as a list of extracted features, \( V = \{f_1, f_2, \dots, f_N\} \). The goal of HAR is to assign an action label \( l \) from a predefined set of actions \( L = \{l^1, l^2, \dots, l^C\} \), where \( C \) is the number of possible actions classes. We aim to find a ``recognition'' transformation $R: V \rightarrow L$ that classifies the video features into possible action categories.

Usually, for a \textit{live} action recognition system, $V$ refers to a buffered history 
\begin{equation}\label{eq:video_feature_buffer}
    V_t = \{f_{t-B+1}, f_{t-B+2}, \dots, f_{t-1}, f_t\},
\end{equation}
where $B$ is the size of the buffer. The recognition algorithm classifies the action class at the current timestamp,
\[\hat{l}_t = R(V_t) \in L.\]
At run time, whenever a new clip $X_{t+1}$, containing the latest frames, becomes ready, the features buffer is updated in a First-In-First-Out manner, with
$V_{t+1} = \{f_{t-B+2}, f_{t-B+3}, \dots, f_{t}, f_{t+1}\}$
and initiates the new inference task, $R(V_{t+1})$.

\begin{figure}[!t]
    \centering
    \includegraphics[width=1\linewidth]{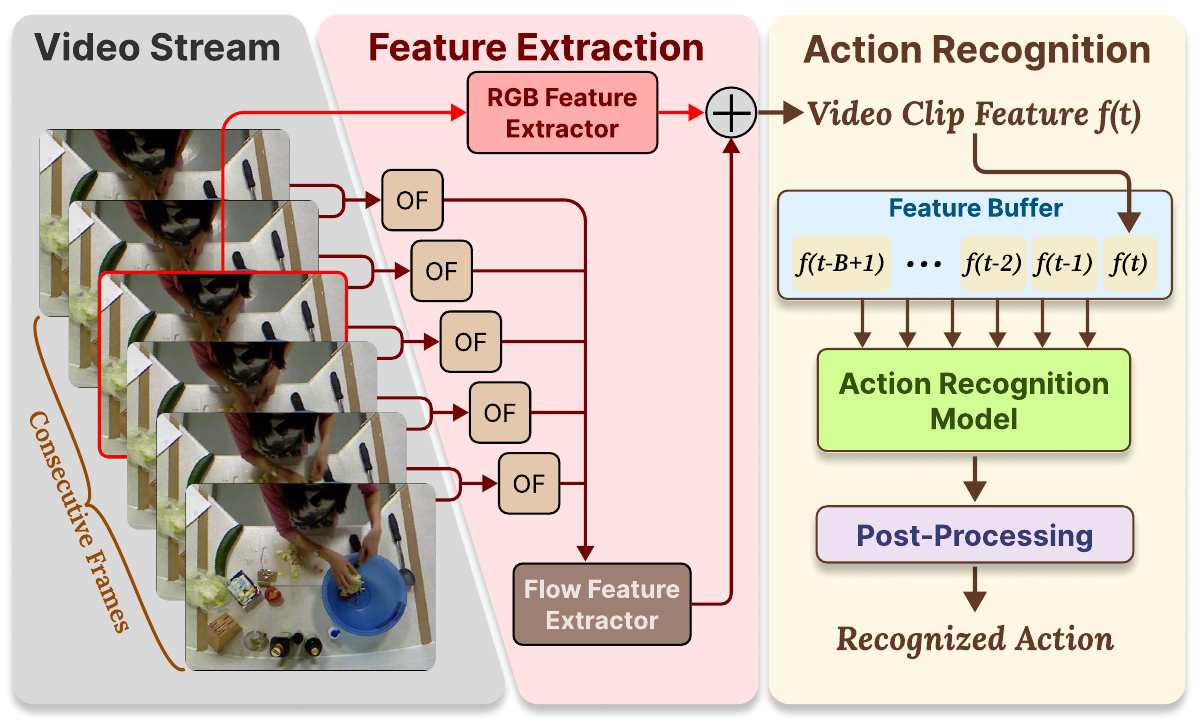}
    \caption{Two-stream video action recognition architecture: video frames are fed into the RGB feature extractor that encodes spatial information. Meanwhile, motion features are extracted by extracting the OFs between frames followed by a flow feature extraction. A buffered history of two-stream features forms the input to the action recognition model to recognize the possible actions, followed by post-processing to refine the output.}
    \label{fig:two_stream_demo}
\end{figure}

\subsection{Soft Real-time Requirement}

Live HAR can be modeled as a periodic task with \textit{soft} real-time requirements. The recognition task initiated by the new clip \( X_t \) is expected to be completed before the next clip \( X_{t+1} \) is available to avoid backlogs and buffer overflow. As a result, the deadline is essentially the duration for $K$ frames to arrive,
\begin{equation}\label{eqn:deadline_formulation}
    T_D = K \times T_{\mathit{INT}},
\end{equation}
where $T_{\mathit{INT}} = {1}/{\mathit{FPS}}$ is the time interval between frames, \ie the inverse of the frame rate ($\mathit{FPS}$) of the video.

\section{Experimental Exploration}

\subsection{HAR Approach}

\label{sec:background:two-stream}

This study specifically focuses on the two-stream feature extractor, illustrated in Fig.~\ref{fig:two_stream_demo}, a foundational building block widely used in state-of-the-art HAR frameworks to extract the features in Eq.~\ref{eq:video_feature_buffer}. A video stream periodically adds video frames into the frame buffer. The \textbf{feature extraction} modules utilize two feature streams to encode the spatial and temporal information separately. Then, the \textbf{action recognition} modules predict the action based on buffered recent features. The prediction results are finally refined with a post-processing filter to smooth out fragmented action predictions. 

Motion feature extraction involves two steps: OF extraction and flow feature extraction. For a $K$-frame video clip, $X_i = \{x_{i,1}, x_{i,2}, \cdots, x_{i,K}\}$, the OF extractor extracts the $K-1$ OFs from consecutive frame pairs:
\begin{equation}\label{eq:flow_feature_K}
\mathit{FLOW}_{i} = \left\{\mathit{flow}_{i,j} | \forall 1\leq j \leq K-1\right\}
\end{equation}
where $\mathit{flow}_{i,j} = \mathit{OF}(x_{i,j}, x_{i,j+1})$. These OFs are then fed to the Flow Feature Extractor to generate motion features.

The RGB feature is extracted from the central frame of the clip, $x_{i,K//2}$, using the RGB feature extractor. We use ResNet50-based 2D CNN feature extractors $E_{\textit{RGB}}$ and $E_{\textit{Flow}}$ for both RGB and flow features. The clip features are extracted by concatenating the outputs from $E_{\textit{RGB}}$ and $E_{\textit{Flow}}$:
\begin{equation}\label{eq:feat_cat}
    f_i = \mathtt{concat}\left(
E_{\textit{RGB}}(x_{i,K//2}), E_{\textit{Flow}}(FLOW_{i})
\right)
\end{equation}
The extracted features are stored in the feature buffer $V_t$ as mentioned in Eq.~\ref{eq:video_feature_buffer}.

% \new{***added LSTR back***}
The action recognition module then recognizes predefined actions from extracted features. To realize live HAR and balance efficiency and performance tradeoff, we employ the Long Short-Term Transformer (LSTR)~\cite{xu2021long} for action recognition. A key characteristic of LSTR is the split of long and short-term memory (buffers), $V = [V_\mathit{long}, V_{\mathit{short}}]$, to handle both short-term and long-range dependencies.  The short-term memory stores the latest features and the long-term memory stores the downsampled older features to reduce the data size and complexity of calculation while preserving the model's capability of capturing action dependencies at a long time scope, making it a practical option for embedded platforms.

\begin{figure}[!t]
    \centering
    \includegraphics[width = 1\linewidth]{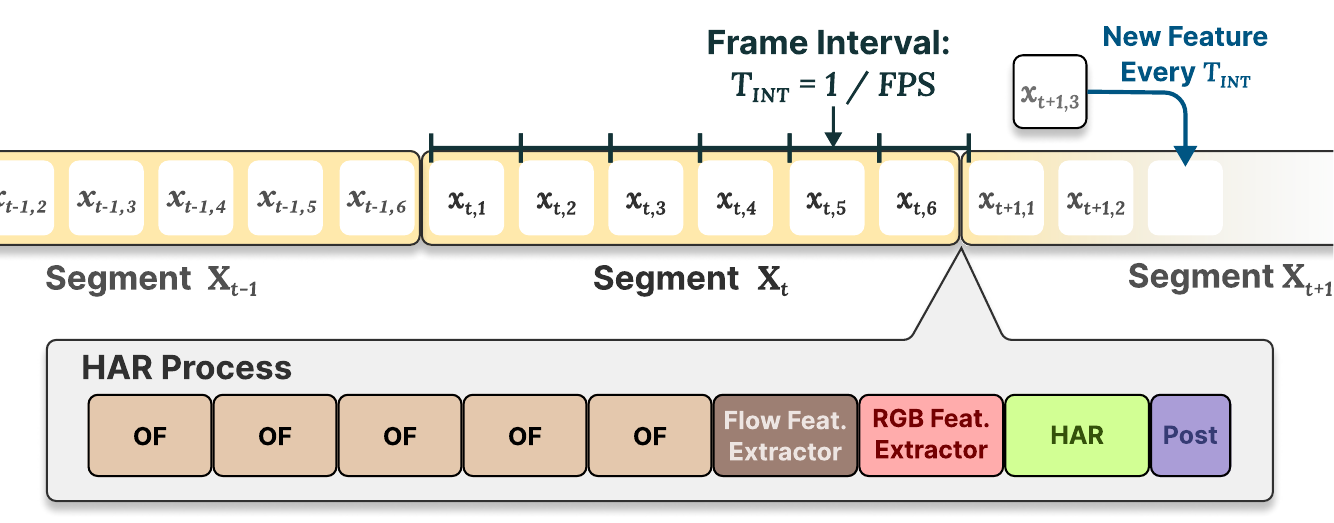}
    \caption{A demonstration of the timing and deadline of a typical two-stream run. The end-to-end latency contains RGB and motion feature extractions followed by recognition and post-processing. Meanwhile, new frames periodically arrive and are stored in the buffer grouped as mini clips.}
    \label{fig:design:deadline}
\end{figure}

To achieve real-time recognition, the system should complete motion feature extraction, RGB feature extraction, recognition, and post-processing by the deadline associated with the frame rate. Fig.~\ref{fig:design:deadline} shows the latency breakdown of a typical two-stream HAR inference process triggered by a new clip. The end-to-end latency of each HAR inference should, on average, be less or equal to the deadline to meet the \textit{soft} real-time requirement.

\subsection{Pipeline Implementation and Workload on Jetson}\label{sec:Implementation_on_embedded_system}
To build a real-time video-based HAR on embedded platforms, directly migrating a state-of-the-art HAR system designed for live recognition seems to be a straightforward option.
In this work, we deploy our live HAR frameworks on a Jetson embedded system\footnote{Jetson Embedded Systems: \url{https://www.nvidia.com/en-us/autonomous-machines/embedded-systems/}}. Specifically, we use a \textit{Nvidia Jetson Xavier NX} (Jetson) embedded AI platform, manufactured by Seeed Studio named as ``reComputer J2021,'' as our testbed. Jetson has a six-core NVIDIA Carmel ARM CPU, a 384-core NVIDIA Volta™ GPU with 48 Tensor Cores, and a unified 8GB RAM shared by CPU and GPU. It runs JetPack 5.1.1, providing a Linux operating system (based on Ubuntu 20.04 with a kernel version of 5.10) and necessary DL dependencies. We set the power mode to 6 (20 W and 2 Core), where our benchmark shows the shortest latency when running HAR inferences, disabled the DVFS governor, and maximized the CPU clocks to 1.9 GHz.

On Jetson, we optimize and run all the neural network components, such as feature extractors and LSTR, with TensorRT. As the TV-L1 OF extractor does not belong to the category of neural networks, we extracted the OFs with GPU-accelerated TV-L1 implementation in OpenCV. The details of embedded system deployment are discussed in Sec.~\ref{sec:embedded_system_deployment}.

We perform the experiments of the HAR system on the 50 Salads dataset~\cite{Stein2013CombiningEA}, a real-world benchmark in action recognition, with 50 top-view salad preparation recordings at 30 FPS. More details of the dataset will be discussed in Sec.~\ref{sec:eval:dataset}.

\subsection{Latency Bottleneck}
\label{sec:background:challenge}
In our experiments, when we sequentially input video frames into the HAR pipeline and make action recognition at the designated frequency (one prediction every 6 frames), we observed a high processing time of about 21 minutes for a video with a duration of 6 minutes and 30 seconds. The lengthy processing time means the original solution is impossible for real-time video applications.

\begin{table}[!t]
\centering
\caption{The latency breakdown of action recognition for the default action recognition pipeline with TV-L1 OF, two feature extractors, and action recognition module (LSTR).}
\begin{tabular}{c|rr}
\specialrule{.8pt}{0pt}{0pt}
\multirow{2}{*}{Module} & \multicolumn{2}{c}{Latency (ms)} \\
                          & \multicolumn{1}{c}{\textit{avg}}             & \multicolumn{1}{c}{\textit{std}}            \\ \specialrule{.8pt}{0pt}{0pt}

TV-L1 $\times$ 5                     & 566.32          & 153.08         \\ 
Flow Feat. Extractor                   & 16.01           & 0.65           \\ 
RGB Feat. Extractor                     & 7.63          & 0.79         \\ 
LSTR                      & 20.22         & 0.46           \\ 
 \specialrule{.8pt}{0pt}{0pt}
\end{tabular}
\label{tab:latency_tvl1}
\end{table}

To better understand the computational bottlenecks in this setup, we conducted a comprehensive latency analysis and presented the breakdown of each component in Table~\ref{tab:latency_tvl1}. Single action recognition is made every six frames, requiring five TV-L1 OF extractions between consecutive frame pairs. Surprisingly, we found that the TV-L1 OF extraction, the core in the motion feature extraction process in numerous state-of-the-art solutions, contributes to $93\%$ of total latency on average on the embedded testbed, even though TV-L1 has already taken advantage of GPU acceleration leading to a speedup over running on CPU. The latency also shows extra variability ranging from 392 ms to 1642 ms with a standard deviation of 153 ms. In the plot, the latency and variability of other modules are trivial. The disproportionate latency and variability of the TV-L1 explains the observed long processing time.

\subsection{Impact of Video Down-sampling}
\label{sec:background:downsample}
It sounds natural to fit the system latency into the actual playback speed of the video for live prediction via frame downsampling. A lower video frame rate reduces the number of HAR predictions and relaxes the deadline for each prediction. Although theoretically appealing, taking a $5\times$ downsampling to 6 FPS only reduces the processing time from 21 minutes to 6 minutes and 42 seconds. The reduction in latency is not proportional to the reduction in frame rate. Further investigation shows that the latency of TV-L1 in each recognition increased, and we assume a direct correlation between latency and the intensity of motion in the video, as downsampling intensifies the motion due to an increased frame interval.

\begin{figure}[!t]
    \centering
    \includegraphics[width = 1\linewidth]{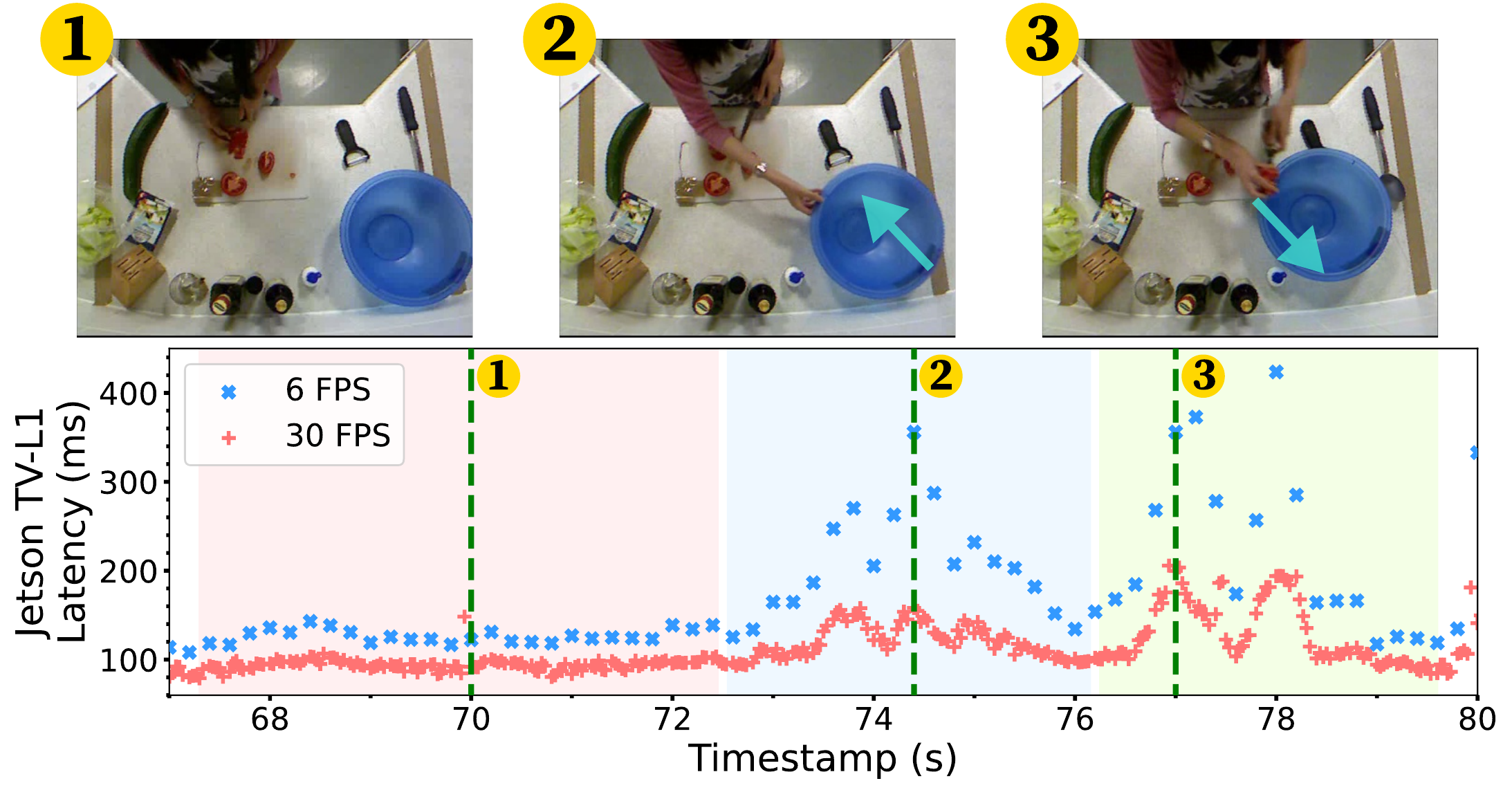}
    \caption{We benchmarked on the Jetson the commonly used TV-L1 OF extractor and visualize latency traces from a video segment, from timestamp 67s to 80s, as indicated on the x-axis. The red markers represent the latency at the original rate (30 FPS), and the blue markers represent a 5x downsampling at 6 FPS, with a sparser marker density. Three cases with different motion patterns are displayed: (1) cutting tomatoes on the chopping board with limited motion on hands. (2) fetching the bowl closer to the chopping board. (3) transferring the diced tomatoes into the bowl. The motion in (2) and (3), \ie the bowl and arms, is greater than (1). The blue arrow visualizes the direction of movement of the objects.}
    \label{fig:latency_trace}
\end{figure}

We verified our assumption by examining the corresponding video content with a visualization of TV-L1's latency patterns in Fig.~\ref{fig:latency_trace}. The blue and green regions in the plot correspond to the video segments with larger movements, which also show spiked latency. When we downsampled the video from 30 FPS (red markers) to 6 FPS (blue markers), the latency significantly increased, especially in the regions of stronger motion. 

We explain the increase in latency with the design of the TV-L1 OF extractor, which defines a pyramid of image scales from coarse to fine to capture the motion of different magnitudes. Large motion displacement may need more pyramid levels to obtain a refined track than a small one, increasing the algorithm's computational cost and latency. The co-occurrence of large (\eg body motion) and small motion (\eg chopping) in the 50 Salads cooking setting explains the unstable latency over time. 

\begin{figure}[!t]
    \centering
    \includegraphics[width = 0.75\linewidth]{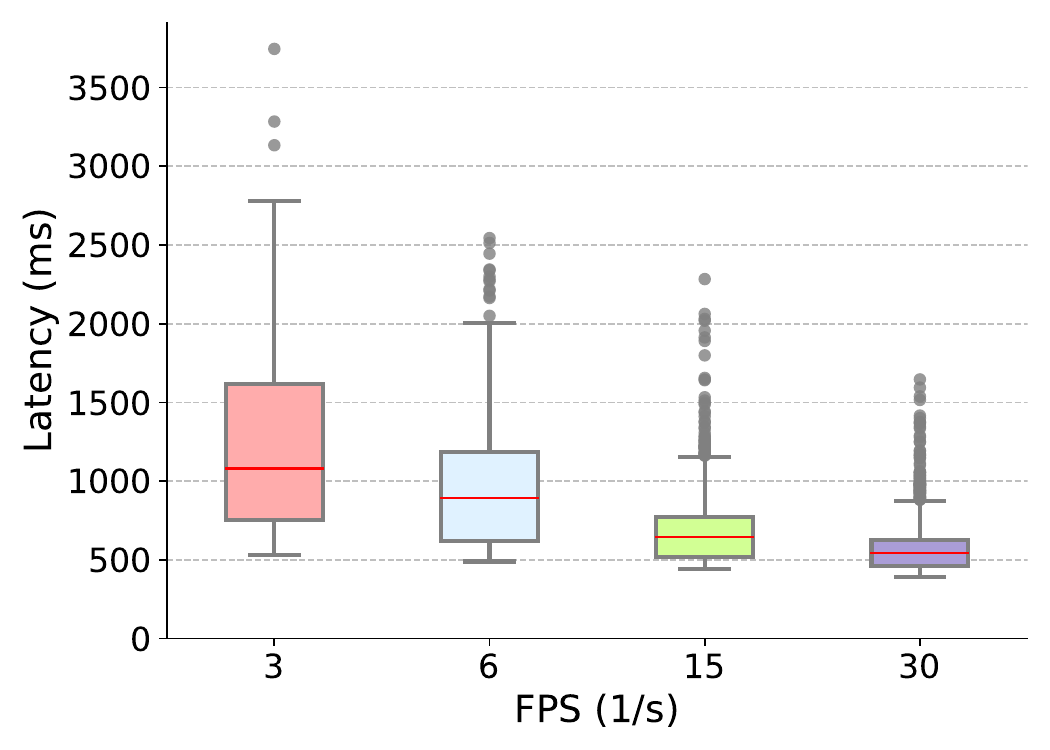}
    \caption{The latency of five TV-L1 OF extractions for video frames at different frame rates. Overall, the latency is the highest at the lowest FPS and the lowest at the highest FPS which indicates a correlation between latency and motion intensity.}
    \label{fig:moti:latancy_tvl1_baseline_2}
\end{figure}

Fig.~\ref{fig:moti:latancy_tvl1_baseline_2} shows a broader view of TV-L1 latency variations with decreasing frame rates. As the inter-frame motion increases, the average and variability of TV-L1 latency also rise incrementally. 

\subsection{Extracing OF with Neural Networks}
\label{sec:background:nueral_network}
To address the large latency and instability caused by the TV-L1 OF extractor, 
we explore neural networks to extract OF as they can take advantage of runtime optimization with neural network inference libraries (e.g., TensorRT), resulting in more stable computations. Among state-of-the-art OF networks, e.g., RAFT~\cite{teed2020raft}, FlowNet 2.0~\cite{ilg2017flownet}, FlowFormer~\cite{huang2022flowformer}, GMFlowNet~\cite{zhao2022global}, RPKNet~\cite{Morimitsu_Zhu_Ji_Yin_2024}, we turn to RAFT for OF extraction due to its efficiency and top-tier OF quality. RAFT is representative of state-of-the-art neural networks for OF extraction that adopt a similar computation procedure including frame feature extraction, cost volume construction with correlations~\cite{teed2020raft} or transformer blocks~\cite{huang2022flowformer}, iterative refinement and upsampling. We select the pre-trained RAFT-Large variant due to its relatively high OF quality in comparison to RAFT-Small. On average, the latency of one RAFT-Large inference is 26.64 ms, with \textit{std} = 2.02 ms. While this method fits within real-time processing constraints, it results in a notable 9.8\% loss in HAR accuracy.  More implementation and evaluation details will be provided in Sec.~\ref{sec:eval}.

While RAFT compromises recognition performance for runtime efficiency, an optimal solution should balance high-quality feature extraction with minimal latency. Our explorations into the second option reveal that redesigning the feature extractor yields exceptional results. The following sections provide the design considerations and implementation specifics of our proposed \proposed.

\section{Integrated Motion Feature Extraction}
While RAFT significantly accelerates  OF extraction compared to TV-L1, inefficiencies persist. As illustrated in Fig.\ref{fig:original_raft_resnet}, the improved two-stream implementation consists of RAFT and a ResNet50-based feature extractor. RAFT includes a context encoder and feature encoders for two consecutive input frames, which are then correlated and processed through a motion encoder, a recurrent neural network (RNN), and upsampling to produce OF vectors. Notably, RAFT needs to run $K-1$ times to generate a stack of OF vectors (Eq.\ref{eq:flow_feature_K}) to form the input to the ResNet50-based feature extractor.

We identified the following weaknesses inherent in such a design:
\begin{itemize}
    \item  A typical first step in neural-network-based OF extractors~\cite{teed2020raft, jiang2021learning, huang2022flowformer} is employing convolutional \textit{Feature Encoders} on the input frames to encode regional information at a reduced resolution. However, this design becomes inefficient when processing consecutive video frames, as one frame will be encoded twice when extracting the OFs with frames before and after.
    % it leads to similar feature encodings to preceding frames.
    \item  Although OF extraction takes the majority of inference latency, it is important to note that OF vectors are only used as intermediate results; they are not used directly by the action recognition model.
    \item  Constructing OF vectors at the original frame resolution leads to unnecessary expansion of data dimensionality, thereby increasing network complexity and latency. 
\end{itemize}

To address these problems, we propose \proposed, an integrated motion feature extractor that uses a single neural network to replace the combination of the OF model and the flow feature extractor and generates the motion features in a single pass.

\subsection{Design of \proposed}
\label{sec:int:design}

\begin{figure}[!t]
    \centering
    \includegraphics[width = 0.8\linewidth]{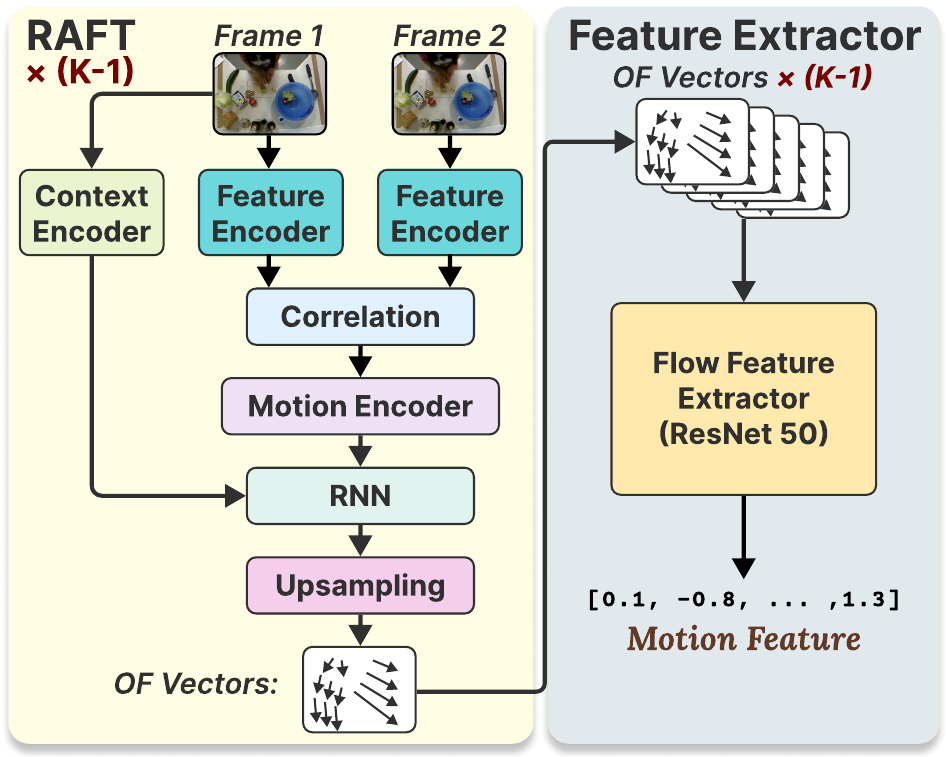}
    \caption{Traditional motion feature extraction combining RAFT as the OF extractor and ResNet50 as the flow feature extractor. RAFT has to execute $K-1$ times to generate the number of OF vectors needed.}
    \label{fig:original_raft_resnet}
\end{figure}

\begin{figure}[!t]
    \centering
    \includegraphics[width = 0.6\linewidth]{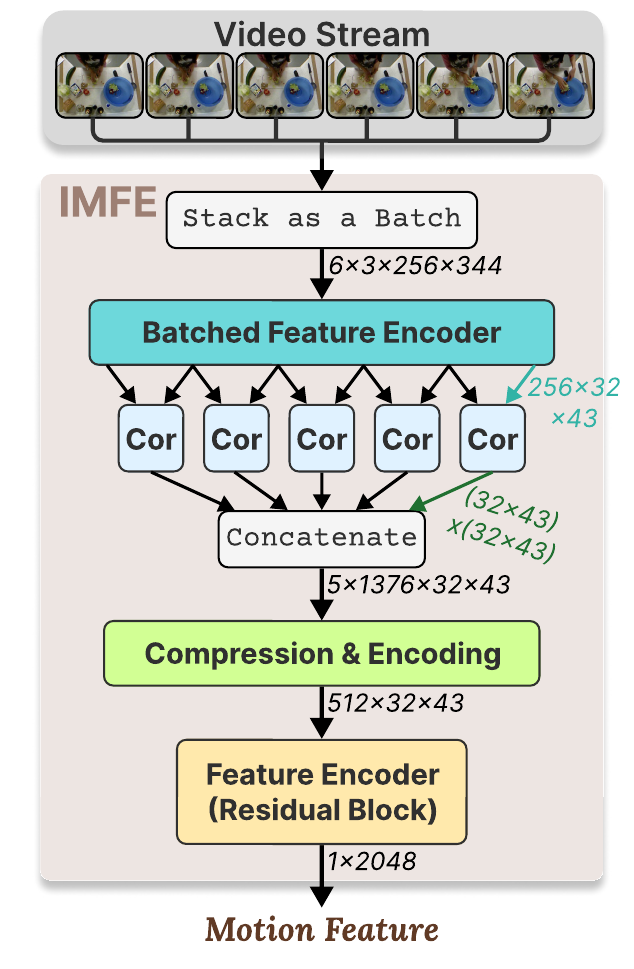}
    \caption{   
    The architecture of \proposed for motion feature extraction. Video frames are stacked as a batch and processed through a Batched Feature Encoder. The consecutive outputs are correlated (Cor) and then concatenated, followed by a Compression and Encoding module and a final Feature Encoder to produce the motion features. The dimensions next to each arrow represent the data's channel, height, and width when the input size is $256\times344$ .
    }
    \label{fig:design_combmodel}
\end{figure}

As shown in Fig.~\ref{fig:original_raft_resnet}, traditional motion feature extraction follows a two-stage process: 1) capturing the correlations between two video frames by extracting OF, and 2) condensing the correlation information into the motion feature vector using the Flow Feature Extractor. Instead of extracting OF as an intermediate input to the Flow Feature Extractor, we propose to extract motion features \textit{directly} from the inter-frame correlations computed at the beginning of the OF extraction process, thereby drastically reducing the complexity and latency of the motion feature extraction process.
The architecture of our novel motion feature extractor, \proposed, is shown in Fig.~\ref{fig:design_combmodel}. It extracts motion features without OF. Furthermore, \proposed takes all $K$ video frames as inputs in one batch. The ``Batched Feature Encoder'' in Fig.~\ref{fig:design_combmodel} eliminates duplicated feature extraction.  

Specifically, for a batch of $K$ consecutive frames represented as $X \in \mathbb{R}^{K \times 3 \times H \times W}$, a feature encoder, consists of 6 residual blocks, produce features maps of frame content downscaled at ${1}/{8}$ of their original size. The feature map is of shape ${K \times D \times \frac{H}{8} \times \frac{W}{8}}$, with $D$ being the dimension of the feature vector representing each position. 
Then, feature-wise correlations between consecutive frames are computed with the correlation modules (the ``Cor'' blocks in Fig.~\ref{fig:design_combmodel}). It generates $K-1$ correlation volumes of shape $\left(\frac{H}{8} \times \frac{W}{8}\right) \times \left(\frac{H}{8} \times \frac{W}{8}\right)$, which is dot-product of vector pairs between consecutive frames to correlate displaced pixel patches. In contrast to prior OF networks that generate multi-level correlation pyramids to detect motion across various scales, our design maintains only the finest resolution level at ${1}/{8}$ scale to reduce complexity. These $K-1$ volumes are then concatenated to consolidate correlation information (the ``\texttt{Concatenate}'' block in Fig.~\ref{fig:design_combmodel}). 
To ensure efficiency in motion feature extraction and address the increased data dimension after concatenation, we employ a Compression and Encoding Module, which consists of five Conv2D layers. The first two layers compress the increased data size resulting from concatenation, significantly reducing data complexity in subsequent processing. The last three layers further extract and condense the motion information into dimensions compatible with the following modules. In the final stage, the Feature Encoder encodes latent information into motion features, utilizing nine residual blocks. The residual modules, fundamental to the ResNet50 architecture, are commonly used for feature extraction in deep learning, leveraging a well-tested architecture to enhance efficiency and effectiveness.
The output dimension of the encoded feature is kept unchanged to preserve structural consistency and ensure seamless integration with subsequent HAR modules in the established frameworks.

\proposed eliminates the need for reconstructing motion vectors by directly encoding the motion and contextual information into motion features. This design, compared with the combination of RAFT and Flow Feature Extractor, bypasses the upsampling of data from the feature dimension to the original image resolution, significantly reducing the complexity of the feature extraction process. As a result, \proposed effectively enhances efficiency and reduces latency in motion feature extraction. While motivated by and evaluated in the context of real-time HAR, \proposed can also be used as the motion feature extractor for other vision applications to improve real-time performance on embedded platforms.

% \subsection{Network Optimization for Accurate Feature Extraction}
\subsection{Training Strategy}

\begin{figure}[!t]
    \centering
    \includegraphics[width = 1\linewidth]{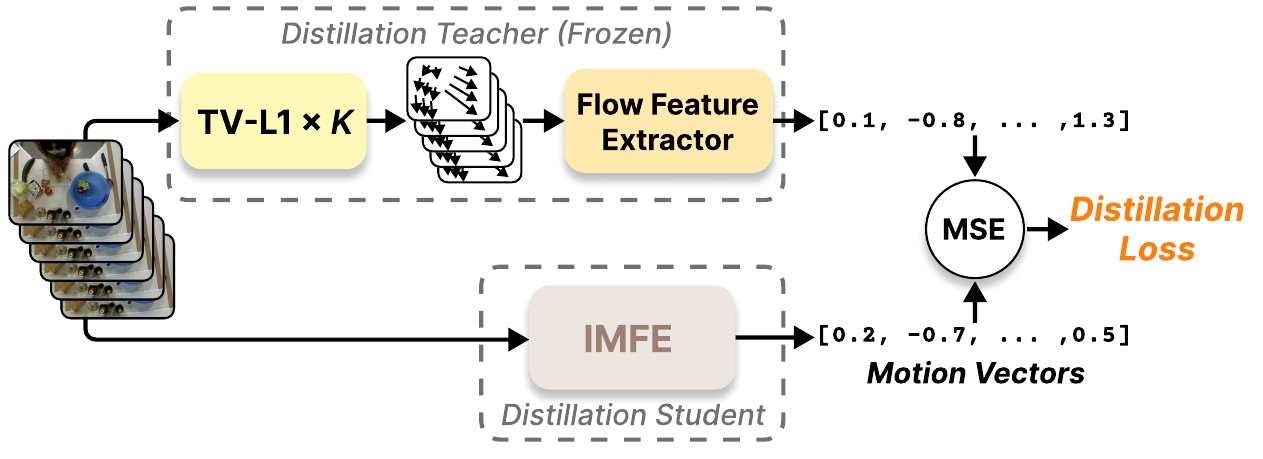}
    \caption{Visualization of the training strategy. Feature extraction with TV-L1 works as the teacher to guide the training of \proposed in the knowledge distillation process. We minimize the MSE between two generated features so that \proposed will generate similar motion features for optimal HAR performance.}
    \label{fig:training}
\end{figure}

A challenge for the new integrated motion feature extractor lies in training \proposed to generate appropriate motion features. 
%However, it is indirect to define a training objective directly to evaluate how well the extracted feature contributes to accurate action recognition. 
One intuitive way to achieve this is to ensure  \proposed produces similar motion features as the one using the standard two-stream method. If we denote \proposed as $E_{\textit{\proposed}}$, the feature extracted is 
$f_{\textit{\proposed}, i} = E_{\textit{\proposed}}\left(X_i\right)$ (The lower path in Fig.~\ref{fig:training}.), and the one from the original flow feature extractor is $f_{\textit{Flow}, i} = E_{\textit{Flow}}\left(\textit{FLOW}_i\right)$ (The upper path in Fig.~\ref{fig:training}.). Our objective is to minimize the difference between $f_{\textit{\proposed}, i} $ and $ E_{\textit{Flow}}\left(\text{TV-L1}\left(X_i\right)\right)$ given any video inputs. 
As a result, we utilize knowledge distillation~\cite{hinton2015distilling} which is proven efficient for lightweight models to learn from large models~\cite{10.5555/3455716.3455856, howard2017mobilenets, sanh2020distilbert, oquab2023dinov2}.
Such consideration not only helps the model learn from the well-established TV-L1 method but also regularizes the feature to encode motion information only. More importantly, during network training, we can take any video clip as input and automatically generate the target feature with supervision from the standard feature extraction method to make the best use of available videos. As in the rightmost part in Fig.~\ref{fig:training}, we use the Mean Square Error (MSE) as the distance metric to quantify the difference between the two extracted features and minimize the following objective,
\[
    \mathcal{L} = \mathbb{E}\left[
        \text{MSE}\Bigl(
            E_{\textit{\proposed}}\left(X\right)
            , 
            E_{\textit{Flow}}(\text{TV-L1}\left(X\right))
        \Bigr)
    \right].
\]

A way to achieve this is to ensure that \proposed should produce similar motion features as the one using the standard TV-L1 method. If we denote \proposed as \(E_{\textit{\proposed}}\), the feature extracted is 
\(f_{\textit{\proposed}, i} = E_{\textit{\proposed}}(X_i)\) (the lower path in Fig.~\ref{fig:training}), and the one from the original flow feature extractor is \(f_{\textit{Flow}, i} = E_{\textit{Flow}}(\textit{FLOW}_i)\) (the upper path in Fig.~\ref{fig:training}). Our objective is to minimize the difference between \(f_{\textit{\proposed}, i}\) and \(E_{\textit{Flow}}(\text{TV-L1}(X_i))\) given any video inputs.

To achieve this, we utilize knowledge distillation~\cite{hinton2015distilling}, which has been proven efficient for lightweight models to learn from larger models~\cite{10.5555/3455716.3455856, howard2017mobilenets, sanh2020distilbert, oquab2023dinov2}. This approach not only helps the model learn from the well-established TV-L1 method but also regularizes the feature to encode motion information exclusively. More importantly, during network training, we can take any video clip as input and automatically generate the target feature with supervision from the standard feature extraction method, maximizing the utility of available videos.

As illustrated in the rightmost part of Fig.~\ref{fig:training}, we use the Mean Square Error (MSE) as the distance metric to quantify the difference between the two extracted features and minimize the following objective:
\[
    \mathcal{L} = \mathbb{E}\left[
        \text{MSE}\Bigl(
            E_{\textit{\proposed}}(X)
            , 
            E_{\textit{Flow}}(\text{TV-L1}(X))
        \Bigr)
    \right].
\]

This loss function helps \proposed learn to generate features that closely mimic those produced by the TV-L1 method, thereby maintaining high accuracy despite the reduced network complexity.

\section{Implementation}
We trained \proposed and the LSTR recognition module in \framework on a DL server and deployed the live HAR pipeline on an embedded system with TensorRT as the optimization technique for efficient inference. This section provides the model training and deployment details and the hardware specifications for the computing hardware we used.
\subsection{Hardware}
We use the same Jetson embedded platform as discussed in Sec.~\ref{sec:Implementation_on_embedded_system}. Model training and export are on a DL workstation with AMD Ryzen Threadripper 3960X 24-Core Processor, 128GB RAM, and dual Nvidia GeForce RTX 3090 GPUs. The server runs Ubuntu 20.04 with a kernel version of 5.15. The ``Distributed Data Parallel'' package is used to facilitate multi-GPU training. We will open source the details of the training environment and code.

\subsection{\framework}
\framework use the same two-stream architecture in Fig.~\ref{fig:two_stream_demo} except with \proposed as the motion feature extractor.
Consistent with prior frameworks, \framework uses the same clip length $K = 6$, meaning that action recognition is made every 6 frames. 
For all feature extraction methods in this work, we keep consistent with the dimensions in the TV-L1+ResNet50 feature extraction with a short-side length of 256 pixels and an aspect ratio of 4:3. 
To meet the requirement of the feature encoding process in RAFT and \proposed that the input shape should be multiples of 8, we round the height and width to the nearest value, \ie $H\times W = 256\times344$. We develop and train all neural networks in PyTorch.

%%%%%%%%%%%%%%%%%%%%%%%%%%%%%%%%%%%%%%%%%%%%%%%%%%%%%%%%%%%%%%%%%%%%%%%%%%%%%%%%%%%%
\subsubsection{\proposed}
We set the dimension of the feature map after the batched feature encoder in \proposed, $D = 256$, as described in Sec.~\ref{sec:int:design}. Consistent with the original flow feature extractor, the motion feature generated by \proposed will be a vector of length 2048.

The training process takes a subset of 4,359 videos from the ActivityNet\cite{7298698} dataset, where 4,059 videos are used for training and the rest 300 for validation. We note that each video is separated into multiple clips and forms training and validation sets of sufficiently large size. The models are trained with a batch size of 16 on each GPU using the AdamW optimizer~\cite{loshchilov2017decoupled} with the default parameter that learning rate~=~0.001. Since the number of training iterations is large, the learning rate is multiplied by 0.9 every 1/5 of the iterations in each epoch. The MSE on the validation set is used to monitor the convergence and over-fitting of the model. We select the checkpoint obtained after 9 training epochs. Although in real-world applications, fine-tuning with the task-specific videos increases the recognition performance at evaluation, we did not fine-tune the model on the 50 Salads dataset based on the following rationale: It allows us to make equitable comparisons with other baseline methods that were not fine-tuned. Without fine-tuning, similar to the feature extraction with TV-L1 and RAFT, \proposed is designed to work as a general model to be easily applied to all HAR models, datasets, and application scenarios. 

%%%%%%%%%%%%%%%%%%%%%%%%%%%%%%%%%%%%%%%%%%%%%%%%%%%%%%%%%%%%%%%%%%%%%%%%%%%%%%%%%%%%

\subsubsection{LSTR}
We universally set the length of short-term memory to $B_\mathit{short} = 80$ features and the long-term memory to 32 features uniformly sampled every fifth feature from the long-term memory of $B_\mathit{long} = 160$ features. We did a grid search on a range of long and short-term memory as in~\cite{xu2021long}. We observed similar accuracy saturation after long-term memory is longer than $16$ and limited effect of the length of short-term memory.  As a result, we selected such parameters with high recognition accuracy at the original frame rate and reasonable model complexity on the embedded system. 
We train the recognition transformer modules in each experiment for 150 epochs, using LSTR's default cross-entropy loss with an additional smooth loss as described in Sec.~3.3 in~\cite{asformer} and employing the Adam optimizer and a cosine learning rate scheduler as in the original work.

\subsubsection{Post-Processing}
As continuity is an inherent property of procedural actions, we incorporate a post-processing mechanism to refine prediction results by resolving the over-segmentation issue. We synthesize two widely-used post-processing approaches: the \textit{boundary regression} approach~\cite{wang2020boundary,Ishikawa2020AlleviatingOE} and \textit{threshold filtering} approach~\cite{10208426,10208681,9866031}. 
The boundary regression builds an ``action boundary barrier'' from action probabilities and unifies the predictions in the segment between consecutive boundaries based on the output of the HAR model, while the threshold filtering filters out the ``noisy" segments with low probability scores or short duration. However, the filtering methods lack semantic information from real action predictions, and the boundary regression approach cannot effectively handle situations with two predicted action boundaries being too close to each other. Thus, we combine the two approaches to mitigate the limitations of either method effectively. In addition, We intentionally set short window sizes with the purpose of minimizing the extra latency introduced and slightly adjust window size for different frame rates.

%%%%%%%%%%%%%%%%%%%%%%%%%%%%%%%%%%%%%%%%%%%%%%%%%%%%%%%%%%%%%%%%%%%%%%%%%%%%%%%%%%%%
\subsection{Embedded System Deployment}\label{sec:embedded_system_deployment}
At deployment, we export each trained model to the Open Neural Network Exchange (ONNX) format for interoperability on the embedded system. Models are optimized and run on the Jetson with TensorRT, an automatic model optimization and acceleration tool for neural network inference. Specifically, we enable mixed precision execution (fp32 and fp16) for all models\footnote{Section ``Reduced Precision'' in the NVIDIA Deep Learning TensorRT Documentation: \url{https://docs.nvidia.com/deeplearning/tensorrt/developer-guide/index.html\#reduced-precision}} and rely on TensorRT to generate the optimized network execution plan. For model layers with parameters exceeding the range of fp16, we manually set the precision of these layers to fp32.

In our current design, the modules are executed sequentially on the Jetson platform. We found that the GPUs are fully utilized during execution, which suggests pipelined execution may not yield significant performance improvement. In Section~\ref{sec:eval:parallel}, we briefly explored parallelization on heterogeneous devices and recognized the potential for integration with existing real-time optimization techniques tailored to various hardware and application scenarios. Detailed investigations are deferred to future work.

\section{Evaluation}
\label{sec:eval}
In this section, we evaluate \framework on the Jetson testbed, focusing on two primary aspects. First, we assess the real-time performance of \framework on the embedded platform by measuring end-to-end latency and deadline miss ratios. Next, we examine the impact of \proposed on action recognition performance, comparing it with current state-of-the-art methods.

%%%%%%%%%%%%%%%%%%%%%%%%%%%%%%%%%%%%%%%%%%%%%%%%%%%%%%%%%%%%%%%%%%%%%%%%%%%%%%%%%%%%
\subsection{Dataset and Frame Rate}
\subsubsection{Evaluation Dataset}
\label{sec:eval:dataset}
We evaluate our HAR solutions on the 50 Salads dataset, a real-world benchmark in action recognition containing 50 top-view camera recordings at 30 FPS of various salad preparation steps, and it aligns well with the practical HAR scenarios. 
We keep consistent with other work using the dataset and use 5-fold cross-validation, averaging the metrics achieved at the last training epoch, to evaluate the quality of features extracted by \proposed and other baselines in differentiating actions. 

\subsubsection{Various Frame Rate}
In practice, HAR systems may not always operate at high frame rates for video inputs and action recognition due to varying scenarios and camera configurations. Additionally, methods with longer latency may struggle to meet real-time constraints when frame rates are high with stricter deadlines as indicated by Eq.~\ref{eqn:deadline_formulation}.
To simulate videos captured under different camera settings and expand our evaluation scope, we downsampled the training and evaluation videos in the 50 Salads dataset to 30, 15, 6, and 3 FPS, which corresponds to $100\%$, $50\%$, $20\%$, and $10\%$ of the original frame rate. Each rate approximately halves the preceding one, with live prediction deadlines of 200 ms, 400 ms, 1000 ms, and 2000 ms, respectively. At each frame rate setting, we re-train all action recognition models for adaptation.

\subsection{Baselines for comparison}
\label{sec:eval:baseline}
To demonstrate the novelty and effectiveness of \proposed's design, we compare our \framework with the following baseline methods. The name of the baselines indicates how the OF vectors are extracted in the original two-stream framework. For all the baseline methods, we follow~\cite{xu2021long} and use the ResNet50-based RGB and flow feature extractors from MMAction2\footnote{MMAction2: \url{https://github.com/open-mmlab/mmaction2}} pre-trained on the ActivityNet dataset. Two feature extractors take an RGB frame and 5 OF frames of size $256 \times 344$, respectively, and encode the spatiotemporal information into a vector of length $2048$.

\begin{itemize}

    \item \textbf{RGB-only:} This variant uses only the RGB features for HAR, eliminating the latency associated with OF extractions and flow feature extraction. It represents the lowest complexity by retaining only the RGB part in Eq.~\ref{eq:feat_cat}.

    \item \textbf{RAFT:} This baseline uses the RAFT-Large model for OF extractions, which is more computationally efficient than TV-L1. We used a pre-trained RAFT available in TorchVision\footnote{RAFT: \url{https://pytorch.org/vision/main/models/raft.html}}. RAFT-Large meets the latency constraints of the frame rate of the dataset (30 FPS) after runtime optimization and significantly improves recognition accuracy over its lightweight variant, RAFT-Small.

    \item \textbf{TV-L1:} This is the standard two-stream method commonly used in HAR solutions including LSTR~\cite{xu2021long, zhao2022testra}. Despite its complexity and instability in TV-L1 OF extractions, which make it unsuitable for real-world deployment, we have still implemented it on both the server and the Jetson and report results at applicable frame rates.

\end{itemize}

\subsection{Latency}

For real-time HAR systems, end-to-end latency reflects the system's ability to make timely predictions and handle high frame rates. To demonstrate \proposed's outstanding efficiency, we measure each module's latency, as shown in Table~\ref{tab:unit_test:of}, for a thorough comparison. For OF modules, we provide the measurements at 30 and 6 FPS to illustrate the neural network's stability to motion magnitude. Finally, we discuss the end-to-end latency (Table~\ref{tab:ete-latency}) of each HAR pipeline on the testbed and discuss the deadline missing ratio under real-time HAR conditions.

\begin{table}[!t]
\centering
\caption{The average latency and the standard deviation of each unique module running on Jetson. GPU acceleration for the TV-L1 OF extractor is used. All neural networks run on Jetson with the TensorRT inference engine and fp32+fp16 mixed precision.}
\begin{tabular}{ccc|rr}
%%%%%%%%%%%%%%%%%%%%%%%%%%%%%%%%%%%%%%%%%%%%%%%%%%%%%%%%%%%%%%%%%%%%%%%%%%%%%%%%%%%%%
\specialrule{.8pt}{0pt}{0pt}
\multirow{2}{*}{Algorithm}    & \multicolumn{2}{c|}{\multirow{2}{*}{Model}} & \multicolumn{2}{c}{Latency (ms)} \\
                              & \multicolumn{2}{c|}{}                           & \multicolumn{1}{c}{\textit{avg}}             & \multicolumn{1}{c}{\textit{std}}             \\ \specialrule{.8pt}{0pt}{0pt}
%%%%%%%%%%%%%%%%%%%%%%%%%%%%%%%%%%%%%%%%%%%%%%%%%%%%%%%%%%%%%%%%%%%%%%%%%%%%%%%%%%%%%
  \multirow{4}{*}{Optical Flow $\times 5$} & \multirow{2}{*}{TV-L1} & 30 FPS & 566.32 & 153.08\\
                                &                        & 6 FPS  & 996.85 & 435.82 \\
                              & \multirow{2}{*}{RAFT}    & 30 FPS & 128.16 & 5.47\\
                              &                          & 6 FPS  & 130.04 & 11.86 \\\hline
%%%%%%%%%%%%%%%%%%%%%%%%%%%%%%%%%%%%%%%%%%%%%%%%%%%%%%%%%%%%%%%%%%%%%%%%%%%%%%%%%%%%%
% \multirow{2}{*}{Feature Ext.} & \multicolumn{2}{c|}{RGB}                        & 28.23 (old)           & 2.44 (old)           \\
%                               & \multicolumn{2}{c|}{Motion}                     & 28.22 (old)            & 5.51 (old)            \\ 
% \multirow{2}{*}{Feature Ext.} & \multicolumn{2}{c|}{RGB}                        & 28.23 (old)           & 2.44 (old)           \\
%                               & \multicolumn{2}{c|}{Motion}                     & 28.22 (old)            & 5.51 (old)            \\ \hline
%%%%%%%%%%%%%%%%%%%%%%%%%%%%%%%%%%%%%%%%%%%%%%%%%%%%%%%%%%%%%%%%%%%%%%%%%%%%%%%%%%%%%
\multirow{2}{*}{Feat. Extraction} & \multirow{2}{*}{ResNet50} & RGB  & 7.11   & 0.89\\
                              &                         &  Flow (RAFT)  & 9.59 & 0.23
                            \\\specialrule{.8pt}{0pt}{0pt}
%%%%%%%%%%%%%%%%%%%%%%%%%%%%%%%%%%%%%%%%%%%%%%%%%%%%%%%%%%%%%%%%%%%%%%%%%%%%%%%%%%%%%
{\textbf{RT-HARE} }         & \multicolumn{2}{c|}{{\textbf{\proposed} (ours)}} & 38.46 & 0.59 \\\specialrule{.8pt}{0pt}{0pt}
%%%%%%%%%%%%%%%%%%%%%%%%%%%%%%%%%%%%%%%%%%%%%%%%%%%%%%%%%%%%%%%%%%%%%%%%%%%%%%%%%%%%%
% \multirow{2}{*}{Action Recog.}  & \multicolumn{2}{c|}{{Two-Stream LSTR}}  & 20.63 & 0.65 \\
%         & \multicolumn{2}{c|}{{Post Processing}} & 0.19 & 0.04 \\
%         \hline
%%%%%%%%%%%%%%%%%%%%%%%%%%%%%%%%%%%%%%%%%%%%%%%%%%%%%%%%%%%%%%%%%%%%%%%%%%%%%%%%%%%%%
\end{tabular}
\label{tab:unit_test:of}
\end{table}

\subsubsection{Latency of Feature Extraction Modules}
Table~\ref{tab:unit_test:of} presents the average latency and standard deviation for each module running on the Jetson platform. The first two rows show the latency for five consecutive TV-L1 OF extractions (OpenCV with GPU acceleration) at two frame rates. The TV-L1 OF extractor exhibits a substantial latency of over 500 ms for five OF extractions, with a high standard deviation. The significant variability in latency aligns with our earlier discussion in Sec.~\ref{sec:background:challenge} on the challenges posed by TV-L1's sensitivity to variations in motion intensity.
In contrast, the RAFT model, executed via TensorRT, demonstrates an advantageous low and stable latency of approximately 130ms for the same task, regardless of the frame rate. RAFT's standard deviation is also considerably lower than that of TV-L1. The fixed architecture of neural networks contributes to more stable latency across different frame rates.

RGB feature extractor's latency is 7.11 ms. The latency for the flow feature extractor differs slightly depending on whether the OF inputs are derived from TV-L1 or RAFT, with RAFT showing 9.59 ms compared to 16.01 ms for TV-L1 as in Table~\ref{tab:latency_tvl1}. Though theoretically, the flow feature extractor should have the same execution time, the observed difference may result from varying CPU-GPU scheduling and device synchronization, as TV-L1 also utilizes the GPU but without the TensorRT library. Regardless, both RGB and flow feature extractors are efficient compared to OF extractions.

Finally, \proposed significantly contributes to the efficiency of the \framework framework, with a latency of just 38.46 ms. As discussed in Sec.~\ref{sec:intro}, the original motion feature extraction process involves both OF extractions and flow feature extraction, which means the combined latency for motion feature extraction using TV-L1 and RAFT and the flow feature extractor at 30 FPS totals approximately 582.33 ms and 137.75 ms, respectively. It follows that \proposed significantly reduces the average motion feature extraction latency by roughly 93\% and 72\% compared to the two-stage process with TV-L1 and RAFT, respectively.

\subsubsection{End-to-End and Real-Time Performance}

\begin{table}[!t]
\centering
\caption{The average end-to-end latency and the standard deviation of each pipeline running on Jetson with video inputs at 30FPS. The upper half shows the results on GPU only and the lower half adopts DLA as an additional processing hardware.}
\begin{tabular}{c|rr}
\specialrule{.8pt}{0pt}{0pt}
\multirow{2}{*}{HAR Pipeline} & \multicolumn{2}{c}{Latency (ms)} \\
                          & \multicolumn{1}{c}{\textit{avg}}             & \multicolumn{1}{c}{\textit{std}}            \\ \specialrule{.8pt}{0pt}{0pt}
TV-L1                     & 614.01          & 139.94         \\ 
TV-L1 (3 FPS)                     & 1473.22          & 623.38         \\ 
RGB-only                  & 24.52           & 1.48           \\ 
RAFT                      & 169.74          & 1.53           \\ 
\textbf{RT-HARE} (ours)             & 68.83           & 2.94           \\ \specialrule{.8pt}{0pt}{0pt}
RAFT + DLA                & 168.69          & 2.97           \\ 
\textbf{RT-HARE} + DLA  (ours)       & 62.98           & 4.18           \\ \specialrule{.8pt}{0pt}{0pt}
\end{tabular}
\label{tab:ete-latency}
\end{table}

As shown in Table~\ref{tab:ete-latency}, we measured the end-to-end latency for each action recognition pipeline on the Jetson platform, including \framework, the original two-stream pipelines using RAFT and TV-L1, and the RGB-only pipeline. The end-to-end latency of each recognition consists of the extraction of two streams of features, action recognition, post-processing, and an additional overhead of less than 5 ms for tasks such as frame resizing and normalization. We benchmark these pipelines on videos recorded at 30 FPS with a corresponding deadline of 200 ms for each prediction.
We also included an additional TV-L1 pipeline at 3 FPS, which is the only configuration on Jetson that meets the corresponding deadline (2000 ms).

The average end-to-end latency of \framework is 68.83 ms, while RAFT and TV-L1 exhibit significantly longer latencies, at 169.74 ms and 614.01 ms, respectively. The end-to-end latency of the RGB-only pipeline is 24.52 ms, with LSTR performing slightly faster due to the absence of motion features.

Interestingly, the end-to-end latency of the RAFT pipeline exhibits a smaller standard deviation compared to five consecutive RAFT OF extractions during unit tests, as shown in Table~\ref{tab:unit_test:of}. Using NVIDIA Nsight Systems (\texttt{nsys}) for profiling on the Jetson platform, we discovered that the synchronization time between the GPU and CPU, along with data transfer, smooths out the end-to-end latency, leading to a lower standard deviation.

We further analyze the deadline miss ratios for each pipeline. The deadline miss ratios for \framework, RAFT, and RGB-only are all 0\% given the 200 ms deadline. In contrast, the TV-L1 pipeline at 30 FPS has a 100\% due to its prohibitively long OF extraction latency. Although the average end-to-end latency of TV-L1 at 3 FPS is much lower than the 2000 ms deadline, it still has a 19.60\% miss ratio due to high latency variance when motion magnitudes change.

\subsection{Inference Parallelization on A Heterogeneous System}
\label{sec:eval:parallel}
In this work, we consider a typical AI-embedded system that uses GPU for all neural network loads. Modern embedded systems are increasingly adopting heterogeneous architectures, incorporating CPUs, GPUs, and Deep Learning Accelerators (DLAs). Task parallelism on different hardware can be utilized to accelerate AI applications. For example, the Jetson Xavier NX includes additional power-efficient DLAs with approximately 36\% of the GPU's performance. The two-stream design is naturally suitable for parallel inference, where we can allocate the less complex RGB stream to the DLA and the motion feature extractor to the GPU. In this subsection, we briefly discuss the system's performance incorporating DLA.

At 30 FPS, \framework, as expected, achieves a similar 63.59\% 5-fold cross-validation accuracy after post-processing with the RGB feature extractor deployed on the DLA and RAFT achieves 59.21\%. The slight but negligible difference in accuracy is due to different TensorRT optimizations and precision settings applied to the different hardware.  As shown in Table~\ref{tab:ete-latency}, through workload parallelization across the GPU and DLA, we decrease the end-to-end latency of \framework from 68.83 ms to 62.98 ms, with a difference of approximately the RGB feature extractor's latency. The end-to-end latency of the RAFT pipeline is also reduced but with a smaller difference from 169.74 ms to 168.69 ms. Both methods exhibit increased latency variance. However, similar parallelization for the RGB stream for the TV-L1 pipeline is less beneficial as its OF extraction process dominates the end-to-end latency with extra instability.

As a future direction, we envision that the \framework can be further accelerated by employing advanced techniques in workload distribution, memory allocation, synchronization, and scheduling. For example, integrating the emerging research in DNN inference scheduling for heterogeneous embedded computing systems~\cite{10.1145/3485730.3485938, 10.1145/3560905.3568520, 10.1145/3581791.3596870} could optimize resource utilization of \framework at a finer granularity.

\subsection{Impact on Activity Recognition Performance}

In this section, we examine the impact of our proposed \proposed on activity recognition performance. We use a set of standard metrics to compare the performance of \framework against various baselines on the targeted dataset, providing a comprehensive evaluation of activity recognition performance.

\begin{table*}[!t]
\centering
\caption{The recognition performances achieved at different frame rates.}
\begin{tabular}{c|c|cccc|cccc|cccc}
\specialrule{.8pt}{0pt}{0pt}
\multirow{2}{*}{} & \multicolumn{1}{c|}{Metric}         & \multicolumn{4}{c|}{Accuracy (\%)} & \multicolumn{4}{c|}{Edit (\%)} & \multicolumn{4}{c}{F1@10 (\%)} \\ \cline{2-14} 
                  &    \multicolumn{1}{c|}{FPS} & 30     & 15     & 6      & 3      & 30    & 15    & 6     & 3     & 30    & 15    & 6     & 3     \\ \specialrule{.8pt}{0pt}{0pt}
RGB-Only          & \multirow{3}{*}{Jetson} &  53.16 &  50.20 & 35.01  & 16.07 & 40.82 & 41.80 & 35.04 & 23.39 & 46.29 & 46.95 & 36.06 & 15.89 \\
RAFT   &                         &  59.24 & 57.09  & 45.56 & 24.72  & 43.44 & 45.34 & 42.56 & 27.43 & 52.20& 52.33 &47.50  & 25.36
\\
\framework   &                         &   63.56  &  59.80 & 45.63  & 21.79  & 44.57 & 49.29 & 43.72 & 26.00 & 54.41 & 56.97 & 48.39 & 22.19
\\ \hline
TV-L1& \multirow{1}{*}{Server} &   69.01 &  64.22 & 48.57  & 21.97  & 52.10 & 49.46 & 43.74 & 20.06 & 61.67 & 59.11 & 49.93 & 22.78 \\
% TV-L1+I3D         &                                      &  73.42 & 70.08 & 62.27  &   39.70   & 59.29 & 56.93 & 58.41 & 36.27     & 68.45 & 64.70 & 63.65 &  43.25   \\
\specialrule{.8pt}{0pt}{0pt}
\end{tabular}\label{tab:training_results_post}
\end{table*}

\subsubsection{Evaluation Metrics}
We use the common metrics for action recognition and action segmentation on the 50 Salads datasets~\cite{Stein2013CombiningEA}, including frame-wise accuracy, F1@k score, and edit~\cite{ding2023temporal}. 
Given a video \( V \) of length \( T \),  a set of predicted frame labels \( \hat{L} = \{\hat{l}_1, \hat{l}_2, \dots, \hat{l}_T\} \) and the corresponding ground truth frame labels \( L = \{l_1, l_2, \dots, l_T\} \).

\textbf{Accuracy}: The frame accuracy is defined as: $A =  \sum_{i=1}^{T} \delta(\hat{l}_i, l_i) / T$, where the function \( \delta(x, y) \) is an indicator function that outputs $1$ if \( x = y \) (i.e., the prediction is correct) and $0$ otherwise. The goal is to evaluate the proportion of the frames correctly classified, and the higher, the better. 

Edit and F1@$k$ metrics additionally focus on evaluating the temporal alignment and segmentation quality of action recognition predictions, beyond mere accuracy. During evaluation, predictions $\hat{L}$ and the ground truth $L$ are refined by merging consecutive identical predictions into action segments that include the action class, start, and end times, $(l, t_{\mathit{start}}, t_{\mathit{end}})$. These segments become \( S_{\hat{L}} = \{\hat{s}_1, \hat{s}_2, \dots, \hat{s}_m\} \) and \( S_{L} = \{s_1, s_2, \dots, s_n\} \), where $\hat{s}_i$ (for $1 \leq i \leq m$) and $s_j$ (for $1 \leq j \leq n$) represent the predicted and ground truth action segments, with \( m \) and \( n \) being the number of predicted and ground truth segments.

\textbf{Edit}: The edit score is calculated as the Levenshtein distance~\cite{lea2016segmental,lea2016learning} between two sequences of actions, i.e., the minimum number of operations needed to transform sequence \( A \) into sequence \( B \).
    \[
    \mathit{Edit}(V) = \mathit{Levenshtein}\left(S_{\hat{L}}, S_{L}\right)
    \]

\textbf{F1@$k$}: The F1@$k$ metric~\cite{Lea_2017_CVPR} evaluates both segment-wise accuracy and segmentation quality. A segment, $\hat{s}_i$, is considered a true positive (TP) if it aligns with a ground truth segment with an intersection-over-union (IoU) exceeding $k/100$, commonly $k=10$, $25$, or $50$. Each ground truth segment can only be matched once. Unmatched prediction segments are false positives (FP), and unmatched ground truth segments are false negatives (FN). With $\textit{precision} = \frac{TP}{TP+FP}$ and $\textit{recall} = \frac{TP}{TP+FN}$, the F1@$k$ score is calculated as:
\begin{gather*}
    \text{F1@}\textit{k} = \frac{2 \times \textit{precision} \times \textit{recall}}{\textit{precision} + \textit{recall}}
\end{gather*}

\subsubsection{HAR Performance}
Table~\ref{tab:training_results_post} shows recognition performances including accuracy, edit, and F1@10, achieved at different frame rates. The table is divided into two sections: the upper section lists methods compatible with the Jetson embedded system, while the lower section shows the performance of more computationally intensive TV-L1 pipeline method on a DL server. We arrange the methods in ascending order of model performance from top to bottom. TV-L1 pipeline can only meet the live prediction deadline at 3 FPS on the Jetson platform as discussed.

We first focus on each method's accuracy as a direct measurement of the HAR quality. The RGB-only pipeline, serving as a lower bound for HAR performance, has the lowest accuracy of 53.16\% at 30 FPS. In contrast, the other pipelines incorporating motion features show significant improvements. RAFT achieves an accuracy of 59.24\%, while TV-L1 achieves the highest accuracy of 69.01\%, indicating that OF quality positively impacts recognition accuracy, with RAFT sacrificing some OF quality for faster inference speed, resulting in lower accuracy.
Our proposed \framework achieves an accuracy of 63.56\%, which lies between RAFT and TV-L1. This result is expected, as \proposed is distilled from its ``teacher'' TV-L1 during optimization and features a lightweight structural design. Despite its simplicity, \framework outperforms RAFT, validating our design assumption that RAFT's OF extraction is redundant and may introduce errors to accumulate. 
As the frame rate decreases, accuracy also declines across all methods, highlighting the significant role frame rate plays in HAR performance. Lower frame rates result in fewer frames for training the recognition model, leading to reduced accuracy. For all methods, higher performance is achieved at higher frame rates, which also shows another benefit of having an efficient HAR framework in real-world applications.

The other evaluation metrics, \ie edit and F1@10, follow the same trend as accuracy. TV-L1 shows the best performance with an edit of 52.10\% and F1@10 of 61.67\% at 30 FPS. The performance of \framework is between that of TV-L1 and RAFT, while RGB-only shows the lowest performance with an edit of 40.82\% and F1@10 of 46.29\%. Unlike accuracy, the best performance in edit and F1@10 is achieved at 15 FPS.
However, lower frame rates result in fewer recognitions, which can mitigate issues with fragmented actions.

Evaluation results demonstrate that \framework achieves an optimal balance between recognition performance and runtime efficiency on modern real-time embedded systems. The novel architectural design of \proposed, specifically tailored for these systems, ensures high HAR accuracy while meeting the real-time requirements of processing live video streams. As modern vision sensors can capture high FPS videos, \framework effectively enables practical HAR applications with real-time video inputs, making real-time video-based HAR and analytics more accessible for embedded systems.

\section{Related Work}

\subsection{HAR Algorithms}
\label{sec:related:video_har}

There has been significant progress in algorithms for video-based HAR. Early works~\cite{baccouche2011sequential, 6165309, 5995496} used a stack of RGB frames as input to recognize human actions with Convolutional Neural Networks (CNNs). The two-stream structure was first proposed in~\cite{simonyan2014two} to recognize actions from both spatial (RGB) and temporal (motion) feature streams. The two-stream design efficiently reduced data dimensionality during feature extraction and, thus, became a popular choice in subsequent HAR algorithms~\cite{ghanem2017activitynet, I3D, farha2019mstcn, li2020mstcn, asformer, Fish2022TwoStreamTA}. 

Real-time applications require HAR to be performed on \textit{live} video streams, where HAR can only utilize video inputs available up to the current timestamp. Live HAR algorithms have been developed using the two-stream architecture recently. OadTR~\cite{Wang_2021_ICCV} recognizes the ongoing action from historical encoded features with a transformer. Long Short-Term Transformer (LSTR)~\cite{xu2021long} splits historical information into long and short memory, sampling the long memory at a lower rate to capture longer time efficiently spans. TeSTra~\cite{zhao2022real} alternatively uses a streaming attention mechanism for recognition. 

As Vision Transformers~\cite{50650} show superior performance in capturing spatio-temporal correlations with its attention mechanism, several transformer-based algorithms have been introduced to perform HAR~\cite{9878733,hussain2022vision,huan2023lightweight} using video frames directly as inputs. However, their computational complexity and memory footprint are prohibitive for embedded platforms due to the high dimensionality of video frame data.

Despite the significant advancement in HAR algorithms, all the aforementioned algorithms were implemented and evaluated on servers. In contrast, our work focuses on real-time performance of live HAR systems on embedded platforms. Henceforth, we adopt the two-stream architecture for its relative efficiency over vision transformers and superior performance over approaches using a single RGB stream. While our implementation of \framework and evaluation employs the transformer-based LSTR as a representative action recognition model, our IMFE motion extractor can be used with other action recognition models or video analytics using motion features with enhanced real-time performance. Furthermore, the insights on achieving real-time performance on embedded platforms may be generalized to other HAR solutions using the two feature streams.

\subsection{HAR on Embedded Platforms}

HAR on embedded platforms remains largely unexplored. Earlier efforts employ primitive algorithms tailored to limited actions. For example, 
Kong et al. \cite{HOGSVM} implemented a fall detection task on embedded systems using histograms of oriented gradients and filtering with a support vector machine classification model. Lightweight CNNs, such as MobileNetV3~\cite{Howard_2019_ICCV} and VGG~\cite{Simonyan2014VeryDC}, have been adopted to extract video features and recognize actions with simple multi-layer perceptron on embedded systems\cite{Kim2021LowCostES,10387836}. However, these works use simplistic algorithms tailored to specific action labels, which cannot be generalized to more complex actions with longer time dependencies. Real-world actions are continuous, interrelated, complex, and often similar, requiring HAR systems to capture these subtleties. To address these challenges, we focus on state-of-the-art HAR algorithms that have superior recognition capabilities but introduce substantial challenges in real-time performance.

\subsection{Real-Time Resource Management for Deep Learning}
There has been significant research on system-level resource management and real-time scheduling of deep learning tasks. For example, LaLaRAND~\cite{9622325} dynamically allocates DNN layers to CPUs or GPUs to enhance schedulability. Prophet~\cite{9984807} optimizes real-time perception for autonomous vehicles by coordinating multiple DNNs on a CPU-GPU architecture. RED~\cite{10405986} employs intermediate deadline assignments to enhance throughput and timeliness in robotic systems under environmental dynamics. The Demand Layering~\cite{9984745} adjusts memory allocation based on current demands for effective resource utilization. RT-LM~\cite{10405961} uses an uncertainty-aware framework to predict output length variability, improving resource allocation for language models.

In this work, we focus on application-level solutions by identifying latency bottlenecks in the HAR pipeline and introducing a novel motion feature extraction approach that leads to a drastic reduction in HAR latency. Our approach is, therefore, complementary to those system-level solutions. An advantage of our application-level approach is that it can be readily deployed on standard platforms (e.g., TensorRT and Linux OS). Future work may explore system-level solutions to further improve the real-time performance of HAR systems.

\section{Conclusion}
In this work, we present an embedded system

This work enables embedded platforms to achieve real-time human activity recognition (HAR) on live videos using advanced deep-learning techniques. We identified that the end-to-end latency in a HAR pipeline is dominated by the optical flow (OF) extractor in the widely used OpenCV library. Although existing deep-learning-based OF extractors can reduce the latency of OF extraction, they result in a considerable decrease in recognition accuracy. To address this, we introduce \proposed, a novel single-shot neural network architecture that extracts motion features without extracting OF, significantly reducing HAR latency with only a moderate impact on recognition accuracy. We developed and evaluated \framework, a real-time HAR system that can perform real-time HAR at least 10 times the video frame rate\footnote{\framework was evaluated at a maximum of 30 FPS with a 200 ms deadline, as limited by the evaluation dataset. However, its end-to-end latency of 69 ms indicates that it can support even higher video frame rates, such as 60 FPS and beyond.} of standard HAR implementations while maintaining high recognition accuracy. Furthermore, \proposed is a highly efficient motion feature extractor that can be integrated into other HAR pipelines and video analytics systems, potentially enabling real-time performance across a wide range of computer vision applications. While \framework has demonstrated substantial improvements in real-time performance, future research could further enhance this by incorporating system-level resource management techniques.

\bibliographystyle{IEEEtran}
\bibliography{IEEEabrv,ref}

\end{document}